\documentclass[conference]{IEEEtran}
\IEEEoverridecommandlockouts
\usepackage{cite}
\usepackage{amsmath,amssymb,amsfonts}
\usepackage{algorithmic}
\usepackage{graphicx}
\usepackage{textcomp}
\usepackage{xcolor}
\usepackage{algorithm}
\newcommand{\beq}{\begin{equation}}
\newcommand{\eeq}{\end{equation}}
\def\BibTeX{{\rm B\kern-.05em{\sc i\kern-.025em b}\kern-.08em
    T\kern-.1667em\lower.7ex\hbox{E}\kern-.125emX}}
\begin{document}

\title{Electric Vehicle Charging Infrastructure Planning: A Scalable Computational Framework \\

\thanks{This work has been funded by the early career development LDRD funding by the Lawrence Berkeley National Laboratory - breakthrough High-performance Computational Solutions to reliable EV-connected Grid Future}
}

\author{\IEEEauthorblockN{Wanshi Hong, Cong Zhang, Cy Chan, Bin Wang}
\IEEEauthorblockA{\textit{Energy Analysis and Environmental Impact Division} \\
\textit{Lawrence Berkeley National Laboratory}\\
wanshihong, congzhang, cychan, wangbin@lbl.gov}}
\maketitle

\begin{abstract}
 The optimal charging infrastructure planning problem over a large geospatial area is challenging due to the increasing network sizes of the transportation system and the electric grid. The coupling between the electric vehicle travel behaviors and charging events is therefore complex. This paper focuses on the demonstration of a scalable computational framework for the electric vehicle charging infrastructure planning over the tightly integrated transportation and electric grid networks. On the transportation side, a charging profile generation strategy is proposed leveraging the EV energy consumption model, trip routing, and charger selection methods. On the grid side, a genetic algorithm is utilized within the optimal power flow program to solve the optimal charger placement problem with integer variables by adaptively evaluating candidate solutions in the current iteration and generating new solutions for the next iterations. 
\end{abstract}

\begin{IEEEkeywords}
Electric vehicle, Charging Infrastructure, Optimization, Transportation, Electricity Grid
\end{IEEEkeywords}

\section{Introduction}
With the increasing penetration of electric vehicles in the transportation system, the charging load is imposing a significant impact on the electric grid. The number of chargers and charger locations will play a deterministic role in EV drivers' route choices and the corresponding load demand onto the local distribution circuits. A systematic approach to quantify the benefits of optimal the placements of charging infrastructure \cite{wang2007optimal,nie2013corridor} to assist stakeholders efficiently distribute the public resources. Agent-based models\cite{sweda2011agent} were developed to optimize electric vehicle charging, hydrogen refueling infrastructure. The National Research Council gave an estimation of the capital investment and operation cost for vehicle charging infrastructure installation\cite{national2013transitions,shareef2016review}, the reliability requirement of the existing distribution grid\cite{zhang2015integrated}. Two objectives \cite{wang2013traffic} have been considered when determining the charging locations, i.e. to minimize voltage fluctuations.  To reduce the operational and investment cost due to charging placements, a stochastic programming approach is proposed in \cite{zhang2018joint} by combining  EV charging behaviors and the on-site photovoltaic along the highways. A number of previous efforts, e.g. \cite{wang_predictive, WANG20171289}, try to optimize the EV charging load considering the driving behaviors and grid constraints. \cite{wencong_7763884} modeled the routing and charging behaviors using mixed-integer programming approaches that are not scalable due to the computational complexities. 

In this paper, we propose a scalable computational framework to optimize the charging infrastructure location selections by coupling the transportation system and the electric grid system, considering a variety of factors that include vehicle energy consumption model, trip routing, and charger selection strategies within the EV trips. First, EV charger selections are simulated within the context of the entire transportation networks and then the charging load is determined and aggregated at the substation level.  For simplicity, we map the aggregated loads to the load buses of IEEE 30-bus standard test case. Second, the resulting charging load profiles are input into the optimal power flow (OPF) program on the grid side to determine the optimal grid nodes for charging infrastructure deployment. We adapted the genetic algorithm to handle the integer variables, which are to denote the charger location selection, and to solve the OPF problem. Compared with previous research, the contributions of this paper are: 1) an integrated infrastructure planning problem, involving complexities of both the transportation side models with the grid side OPF problem; 2) The charging profile is generated from the vehicle dynamic model, which can be extended to cover various kinds of electric vehicles; 3) a novel computational framework with an outer-loop driven by the genetic algorithm to combine grid and transportation models. Note that in this study, we aim to determine optimal placement of charger aggregator, instead of individual chargers. 
\begin{figure}[h]
\vspace{-0.1in}
    \centering
    \includegraphics[width = 1.0\linewidth,trim=0 0 0 0,clip]{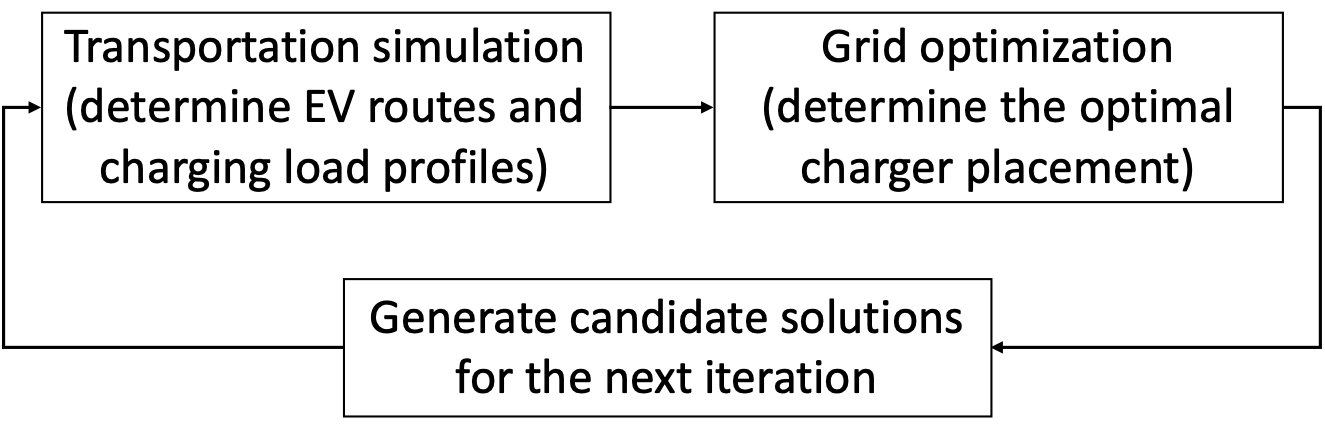}
    \caption{A computational framework with outer-loop driven by genetic algorithms.}
    \label{fig:outer_loop}
\end{figure}
\vspace{-0.2in}

The paper is organized as follows: In section II, we illustrate the EV load generation models within the transportation system context, including the EV energy consumption model, routing method and charger selection logic. In section III, the charger placement is optimized by iteratively solving the OPF problem using the modified genetic algorithm. Finally, we conclude the paper in section IV with potential future work.

\section{EV Charging Load in the Transportation System}

\begin{figure}[h]
    \centering
    \includegraphics[width = 0.8\linewidth]{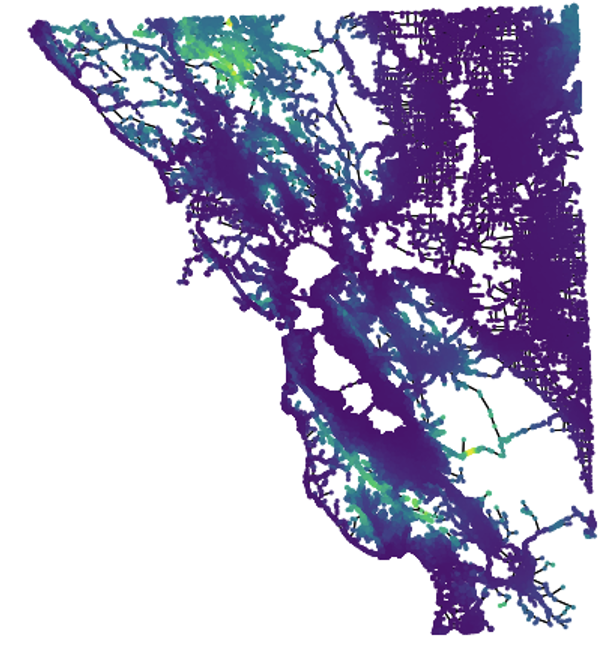}
    \caption{Transportation network in the San Francisco bay area with nodal elevation data.}
    \label{fig:Bayarea}
\end{figure}

In this paper, we simulated the EV charger selection logic within the San Francisco bay area transportation network and determined the corresponding charging load profiles of the selected sites. In the region, there are 24 million trips per day, covering 455 thousand nodes and 1 million road links. For each node, the latitude and longitude info is included. Fig. \ref{fig:Bayarea}  is the visualization of the  network with elevation. Specific modules are discussed as follows.
\subsection{EV energy consumption model}
The scenario consists of 24 million trips with a fraction of them as EV trips. EV energy models are developed to calculate the energy consumption. We synthesized a population of the most popular electric vehicles (EV), including the Tesla Model 3, Nissan leaf, BMW i3 and Chevrolet Bolt. Based on the vehicle specifications, aerodynamic coefficient, front area, mass, battery capacity, which are incorporated within the vehicle dynamic resistance formula\cite{yang2020real}, the instantaneous tractive power can be calculated using the following equations.
\beq\label{eq:vehicleforce1}
F_t = F_f +F_i + F_w+ F_j
\eeq
\beq\label{eq:vehicleforce2}
\frac{T_{tq}i_T\eta}{r}=mgcos\theta\cdot f{}+mgsin\theta+\frac{C_DAu^2}{21.15}+\delta m\frac{du}{dt}
\eeq
\beq\label{eq:vehicleforce3}
P=F_t\cdot u
\eeq
where $F_{t}$ is tractive force, $F_f$ is the fraction resistance,  $F_i$ is the road load resistance force caused by road inclinations,  $F_w$ is the aerodynamic resistance,  $F_j$ is the acceleration resistance, $T_{tq}$ is the torque, $i_T$ is transmission ratio, $\eta$ is powertrain efficiency, $r$ is the radius of wheel, $m$ is vehicle mass, $g$ is the universal gravitational Constant vehicle weight, $\theta$ is the road slope, $f$ is the road fraction coefficient, $i$ is road angle, $C_D$ is vehicle aerodynamic coefficient, $A$ is vehicle front area, $u$ is vehicle speed (km/h), $\delta$ is mass coefficient, $\frac{du}{dt}$ is acceleration, $P$ is tractive power.

\subsection{EV trip routing with charger selection strategy }
With the energy consumption load, the remaining state of charge (SoC) can be calculated throughout the trip. The charger selection depends on the trip distance, trip origin/destination location, and the SoC. Table 1 shows the routing and charger selection strategy during the vehicles driving. Note again, charger here refers to the charger aggregator, where multiple chargers can be deployed. 

\begin{algorithm}
    \caption{EV routing and charger selection strategy}
  \begin{algorithmic}[1]
      \STATE Assign vehicle models to trips
      \STATE Read the vehicle parameters
      \STATE Obtain the trip origin (O node) and destination (D node)
      \STATE Route the trip using shortest-path algorithms
      \STATE Calculate the energy consumption for the selected trip/vehicle 
      \STATE Determine the charger selection and charging operation, i.e. determine charging power profile
      \begin{itemize}
      \item \textbf{IF} remaining SOC $>$ energy consumption on the trip
      
          \textbf{ACTION 1:} No charging is needed
      \item \textbf{ELIF} remaining SOC $>$ 80$\%$ of the energy consumption on the trip
      
          \textbf{ACTION 2}: Find the best charging station (M node) on the trip 
      \item \textbf{ELSE}: (Not able to finish the majority of trip)
      
          \textbf{ACTION 3}: Charging the vehicle at origin
      \end{itemize}
  \STATE \textbf{Output} Charging profile of each charging station  
  \end{algorithmic}
\end{algorithm}

Note that the python-based implementation utilizes Networkx for shortest-path routing. Step 6 indicates the charging selection process where 3 operations can be determined: no charging, charging at origin place and charging during the trip.
\begin{itemize}
    \item \textit{No charging:} It applies to the majority of the trips when the vehicles have enough energy to finish the whole trips.
    \item \textit{Charging at origin place:} It works for the vehicle that cannot finish the majority (factor = 80$\%$) of the next trip. This factor can be adjusted depending on the trip length. 
    \item \textit{Charging during the trip:}  It applies to the vehicle that has enough energy to finish the larger portion (factor = 80$\%$) of the trip, but not the whole trip.
\end{itemize}

\begin{figure}[h]
    \centering
    \includegraphics[width = 1\linewidth]{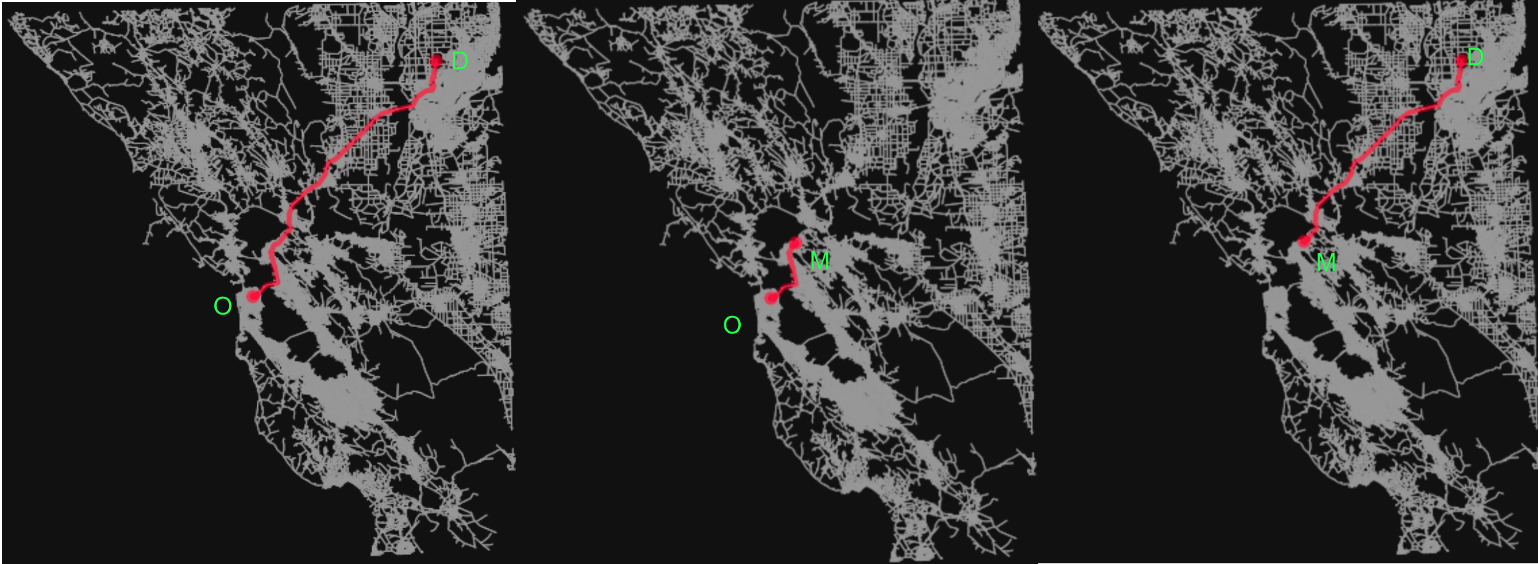}
    \caption{Trip demonstration (a) whole trip(O node to D node) (b) subtrip 1(O node to M node)  (c) subtrip 2(M node to D node).}
    \label{fig:tripdemonstrations}
\end{figure}

In order to search for the best charging station during the trip (along the planned route), two processes are used as follows:.
\begin{itemize}
    \item \textit{Process 1:} Filter all potential charging stations and find the first $N$ nearest charging stations closest to the route. Fig. \ref{fig:Filter} shows the algorithm.
\begin{figure}[h]
    \centering
    \includegraphics[width = 0.8\linewidth]{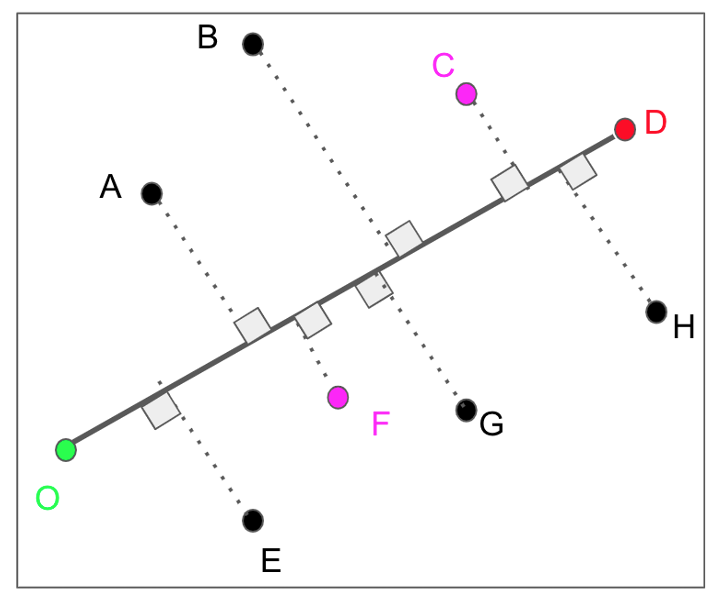}
    \caption{Filter the top N nearest stations (N=20, demo where C and F are selected).}
    \label{fig:Filter}
\end{figure}
    \item \textit{Process 2:} Find the nearest charging station among the stations selected above. Fig. \ref{fig:tripdemonstrations} shows the routing result from node O to D. M is the selected charging station.
\end{itemize}

\begin{figure}[h]
    \centering
    \includegraphics[width = 1\linewidth]{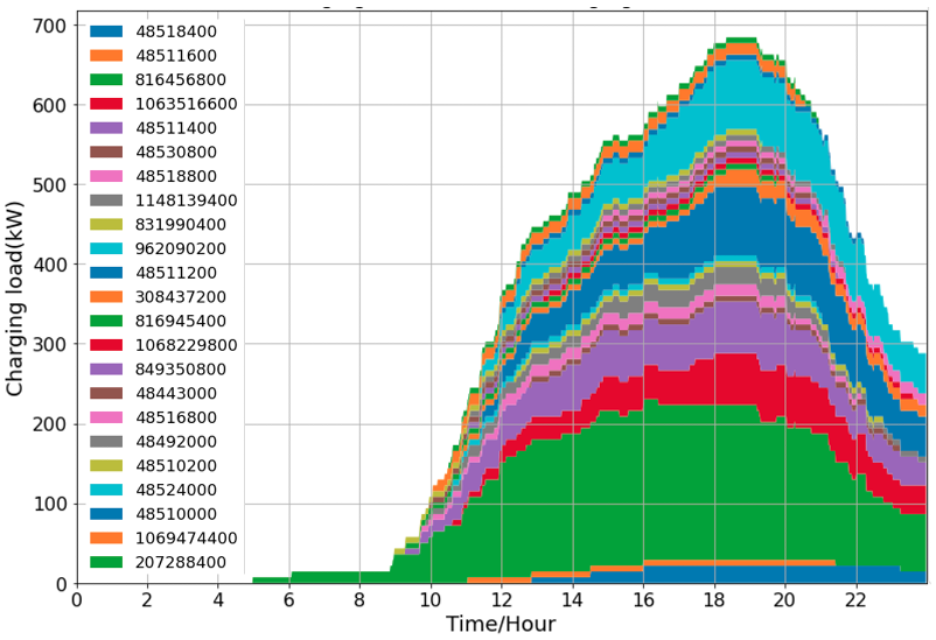}
    \caption{Charging load profiles of different charging stations.}
    \label{fig:chargingloadatchargingstations}
\end{figure}

\subsection{Charging load generation}
Fig \ref{fig:chargingloadatchargingstations} shows the stack charging load profiles of the charging stations. In this study, we simulated 100 light-duty vehicles, and the charging power is assumed level 2(7.2 kW). The time resolution is 1 minute. The resulting charging load, after being aggregated to the corresponding substation nodes, can act as the additional load to the grid system.

\section{Grid-Side Optimal EV Charging Station Placement}\label{sec:grid}
In this section, the optimal charger placement to reduce power generation cost. First, the optimization problem is formulated and the methodology to solve the problem is introduced. Then, a synthetic simulation study based on IEEE 30 bus test case is presented.

\subsection{Optimization problem formulation}\label{subsec:grid-prblm}
From the grid-side planning point of view, the objectives of EV charging station selection are to achieve minimum power generation cost and obtain the optimal EV charging load distribution. To meet the above objectives, we formulate an OPF problem for optimal charger placement as well as the generator operation strategy. For this study, we map the charging station network to a transmission network, where we assume the aggregator of charging stations is considered as an additional load to be added to a proper bus node in the network.

\medskip
\textbf{Cost function}. The OPF problem can be described as a minimizing the total fuel cost of all the committed power plants subject to some constraints. The cost function is given as follows:
\beq\label{eq:gencost}
J =\sum_{t=1}^{T} \sum_{i=1}^N\phi_i(P_{i}^g(t))
\eeq
where $\phi_{i}$ is the fuel cost function of the $i$th generator, $N$ is the number of generator, and $P_{i}^g(t)$ is the active power of the $i$th generator at time step $t$, where $t=1,\dots,T$. The generator cost $\phi_i$ can be expressed as a continuous quadratic equation, indicating the variation of fuel cost ($\$$) with generator power (MW), as follows:
\beq\label{eq:quadcost}
\phi_i(P_{i}^g(t)) = a_iP_{i}^{g2}(t)+b_iP_{i}^g(t)+c_i.
\eeq

\medskip
\textbf{Power flow equations}. The above OPF problem is built based on a power grid modeled by the triple $\{\mathcal{N,G},Y\}$ where $\mathcal{N}=\{1,\dots,n\}$ is the set of buses, $\mathcal{G}\subseteq\mathcal{N}$ is the set of generators with $N$ generators inside. The status of each bus $i$ is represented by its voltage $V_i=|V_i|e^{\mathbf{j}\theta_i}$, where $|V_i|$ is the voltage magnitude, $\theta_i$ is the phase angle at bus $i$, and $Y=G+\mathbf{j}B\in C^{n*n}$is the bus admittance matrix. The OPF problem is subjected with the power flow constraints:
\begin{align}
    P_i = \sum_{k\in \mathcal{N}}^n|V_i||V_k|(G_{ik}\cos(\theta_i-\theta_k)+B_{ik}\sin(\theta_i-\theta_k))\label{eq:pf1}\\
    Q_i = \sum_{k\in \mathcal{N}}^n|V_i||V_k|(G_{ik}\sin(\theta_i-\theta_k)-B_{ik}\cos(\theta_i-\theta_k)),\label{eq:pf2}
\end{align}
where we assume bus $i=1$ as the slack bus with $\theta_1=0$. For simplicity, we further assume there is only one generator at each bus, the apparent power at bus $i$ is given by:
\beq\label{eq:appwr1}
S_i = P_i + \mathbf{j}Q_i = \left\{  
             \begin{array}{ll}  
             (P_i^g-P_i^d)+\mathbf{j}(Q_i^g-Q_i^d), & i\in \mathcal{G}, \\  
             -P_i^d-\mathbf{j}Q_i^d, & i\notin \mathcal{G},
             \end{array}  
\right. 
\eeq
where $P_i^g$ and $Q_i^g$ are the generator active and reactive power at generator nodes, $P_i^d$ and $Q_i^d$ are the uncontrollable power sinks at all the buses.

\medskip
\textbf{EV load distribution}. To model EV charger stations to the OPF problem, we assume there is a charger station set $\mathcal{L}$ with $M$ charger stations to be added to the power system, and each charger station can arbitrarily be added to any buses in $\mathcal{N}$. The EV load can be modeled to equation (\ref{eq:appwr1}) at bus $i$ as:
\beq\label{eq:appwr1}
S_i = \left\{  
             \begin{array}{ll}  
             (P_i^g-P_i^d-\sum_{\substack{j\in \mathcal{L}\\L_{j}=i}}P_j^{EV})+\mathbf{j}(Q_i^g-Q_i^d), & i\in \mathcal{G}, \\  
             -P_i^d-\sum_{\substack{j\in \mathcal{L}\\L_{j}=i}}P_j^{EV}-\mathbf{j}Q_i^d, & i\notin \mathcal{G},
             \end{array}  
\right. 
\eeq
where $P_{j}^{EV}$ is the EV charging load at $j$th charging station, $j = 1,\dots,M$, and $L_{j}$ indicates the charging station location $L = [L_{1},\dots,L_{M}]^T$. For example, $L_2=5$ and $P_2^{EV}=15MW$ indicates the second EV charging station is placed on bus node \#5, and the EV load at the second charging station is 15MW.

\medskip
\textbf{Decision variables}. The decision variable $x$ for the optimization problem is:
\beq\label{eq:dec-var}
x = [P_{g}^T,\  Q_{g}^T,\  L^T]^T,\ \ x\in\mathcal{X},
\eeq
where $P_{g} = [P_{2}^g,\dots,P_{N}^g]^T$ is the generator active power, $Q_{g} = [Q_{2}^g,\dots,Q_{N}^g]^T$ is the generator reactive power $L = [L_{1},\dots,L_{M}]^T$ are the EV charger station locations which are integer variables, $\mathcal{X}$ is the bound for the decision variables including generator power limit and charger location limit. 

One can see that the above formulated OPF problem is a mixed-integer programming problem. To handle the integer decision variables, in this study we propose to use mixed-integer genetic algorithm to solve (\ref{eq:gencost}) - (\ref{eq:dec-var}).

\subsection{Methodology: mixed-integer genetic Algorithms}\label{subsec:grid-MIGA}

Genetic Algorithms (GAs) are optimization algorithms based on the mechanics of natural selection and genetics \cite{bakirtzis2002optimal}. In genetic algorithms, a population of candidate solutions (\textit{chromosomes}) to an optimization problem is evolved toward better solutions using a set of genetic operators:
\begin{itemize}
    \item \textit{Selection}. GAs use selection techniques to select better individuals from the population to be inserted into the mating pool, which is then used to generate new offspring. The selection rule used in our approach is a stochastic universal sampling (SUS) selection method \cite{pencheva2009modelling}.
    \item \textit{Crossover}. A crossover operator is used for the information exchange between mating individuals. The chromosomes of the two parents are combined to form new offspring which inherit information in parent chromosomes. Laplace crossover is used in this study \cite{deep2009real}.
    \item \textit{Mutation}. A mutation operator is used to inject new information. The mutation alters one or more gene values in offspring to generate new characteristics different from the parent population. Power mutation is used in this study \cite{deep2007new}.
\end{itemize}

\textbf{Truncation procedure for integer restrictions \cite{deep2009real}}. To ensure the integer restrictions are satisfied after crossover and mutation operations, we apply the following truncation procedure, where $x_i$ is truncated to the integer value $\bar x_i$ by the rule:
\beq
\bar x_i = \left\{  
             \begin{array}{ll}  
             \lfloor x_i \rfloor & r\in[0,0.5] \\  
             \lceil x_i \rceil & r\in(0.5,1]
             \end{array}  
\right. 
\eeq
where $r$ is a uniformly distributed random number between 0 and 1.

The above algorithm can be summarized as follows:
\begin{algorithm}
    \caption{Mixed-Integer Genetic Algorithm}
  \begin{algorithmic}[1]

    \STATE \textbf{Initialization} Gen = 1, random initial population
    \WHILE{Gen$\leq$MAXGEN}
      \STATE Fitness function (FF) (\ref{eq:gencost}) - (\ref{eq:dec-var}) evaluation
      \STATE Parent selection
      \STATE Crossover
      \STATE Mutation
      \STATE Truncation procedure
      \STATE Gen = Gen + 1
    \ENDWHILE
  \STATE \textbf{Output} Best individual from FF evaluation    
  \end{algorithmic}
\end{algorithm}

\subsection{Simulation study}\label{subsec:gird-sim}
\begin{figure}
    \centering
    \includegraphics[width=\linewidth]{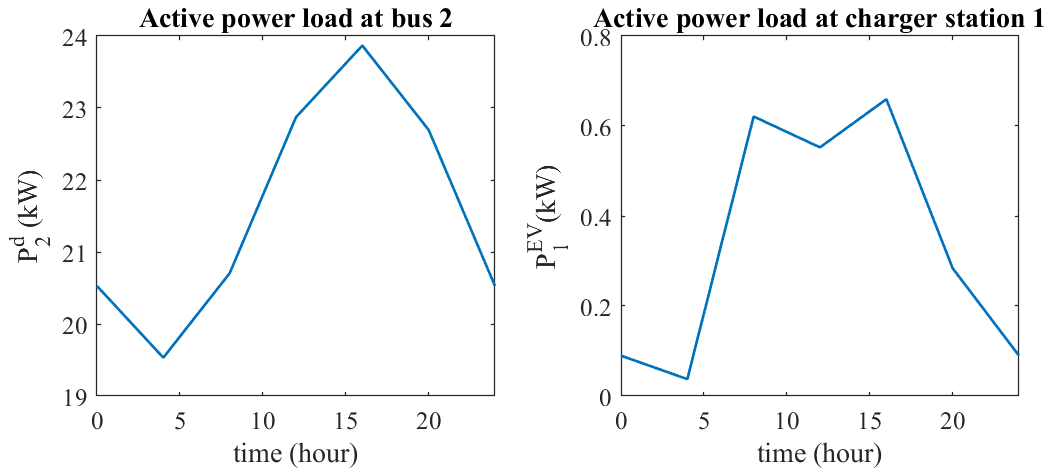}
    \caption{Load profile.}
    \label{fig:load}
\end{figure}

The above algorithm is tested on an IEEE 30 bus test case. With 10 EV charging stations to be assigned. Fig. \ref{fig:load} shows the daily load profile at bus 2 and EV charger station 1 as an example. For the GA algorithm parameters, we choose maximum iteration number MAXGEN=400, crossover probability $P_c=0.9$, mutation probability $P_m=0.1$, size of the population $n_{pop}=2000$.

\begin{figure}
    \centering
    \includegraphics[width=0.8\linewidth]{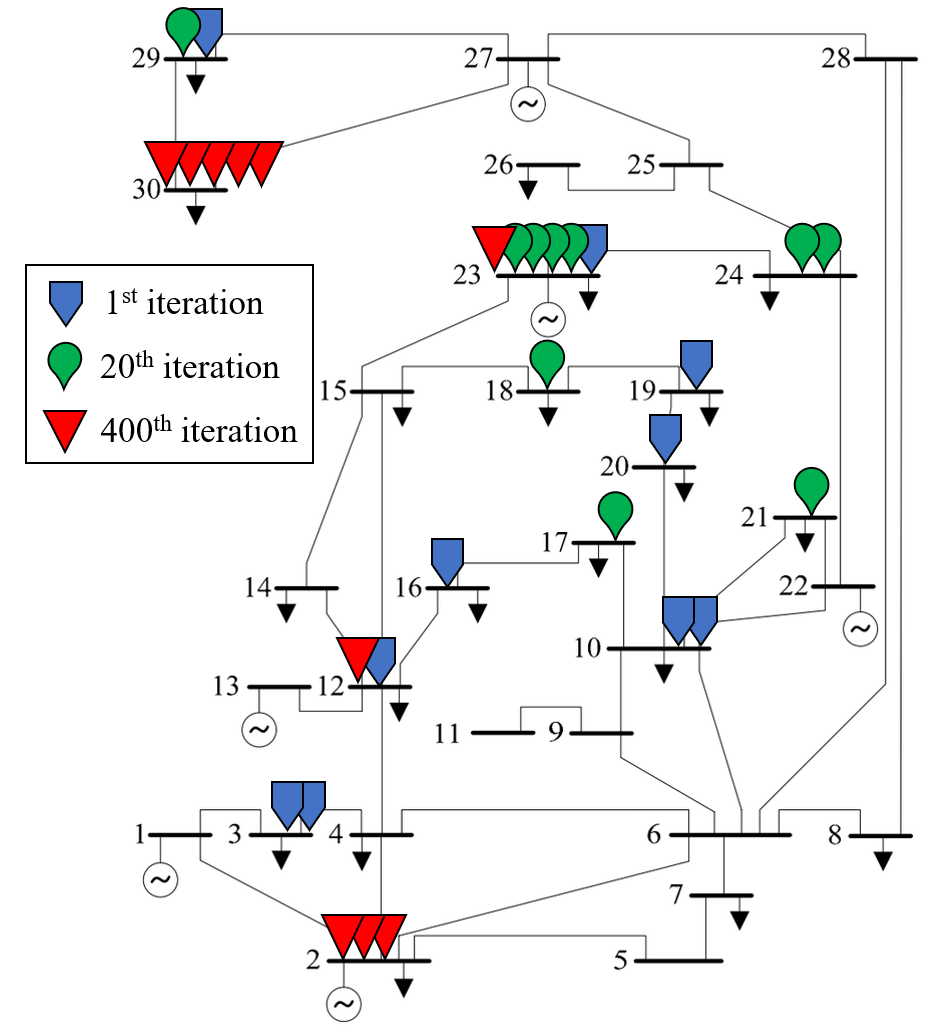}
    \caption{Charger location.}
    \label{fig:charging1}
\end{figure}

\begin{figure}
    \centering
    \includegraphics[width = 0.8\linewidth,trim=0 0 0 10,clip]{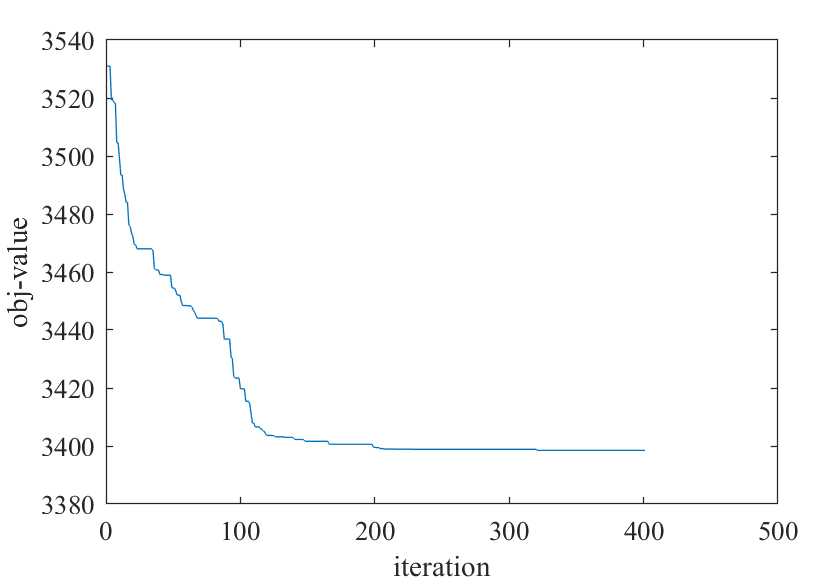}
    \caption{Mixed-integer GA performance.}
    \label{fig:charging2}
\end{figure}

The optimization results are shown in Fig. \ref{fig:charging1} - \ref{fig:charging2}. Fig. \ref{fig:charging1} shows how the charging station is planned at the 1st, 20th, 400th iteration, in blue, green, red pins, respectively. Fig. \ref{fig:charging2} shows the objective value change by iterations. 


\textbf{Discussion}: Based on the optimization results shown in \ref{fig:charging1} - \ref{fig:charging2}, one can see that the optimization problem is converging after 100 iterations, and the optimal EV charger stations are located at bus 2, bus 12, and bus 30. This result physically makes sense as the charger station location is selected close to the generator with a larger capacity or a lower load demand. Combined with the nearest charger station selection in Section II-B, the proposed method can be used to find the best charger stations which are cost-effective from the grid point of view.

\section{Conclusion and Future Work}
In this paper, we developed a computational framework to determine the optimal placement of the EV charging infrastructure by coupling the large-scale transportation and grid system. The simulation study shows some preliminary results of the EV charger placement given EV charging load profiles as inputs. However, the feedback from grid optimization to transportation operation has not been fully enabled within each iteration. The future work is to extend the current framework to support a fully iterative and automatic process that dynamically considers the charging station placement, routing and charger selection process on a real-world grid network.

\section{Acknowledgement}
 This research has been supported by the ongoing early career LDRD project,  "Heavy High-performance Computational Solutions to Secure a Reliable EV-connected Grid Future", at Lawrence Berkeley National Lab.

\bibliographystyle{./bibliography/IEEEtran}
\bibliography{./bibliography/IEEEabrv,./bibliography/IEEEexample}

\begin{thebibliography}{10}
\providecommand{\url}[1]{#1}
\csname url@samestyle\endcsname
\providecommand{\newblock}{\relax}
\providecommand{\bibinfo}[2]{#2}
\providecommand{\BIBentrySTDinterwordspacing}{\spaceskip=0pt\relax}
\providecommand{\BIBentryALTinterwordstretchfactor}{4}
\providecommand{\BIBentryALTinterwordspacing}{\spaceskip=\fontdimen2\font plus
\BIBentryALTinterwordstretchfactor\fontdimen3\font minus
  \fontdimen4\font\relax}
\providecommand{\BIBforeignlanguage}[2]{{%
\expandafter\ifx\csname l@#1\endcsname\relax
\typeout{** WARNING: IEEEtran.bst: No hyphenation pattern has been}%
\typeout{** loaded for the language `#1'. Using the pattern for}%
\typeout{** the default language instead.}%
\else
\language=\csname l@#1\endcsname
\fi
#2}}
\providecommand{\BIBdecl}{\relax}
\BIBdecl

\bibitem{wang2007optimal}
Y.-W. Wang, ``An optimal location choice model for recreation-oriented scooter
  recharge stations,'' \emph{Transportation Research Part D: Transport and
  Environment}, vol.~12, no.~3, pp. 231--237, 2007.

\bibitem{nie2013corridor}
Y.~M. Nie and M.~Ghamami, ``A corridor-centric approach to planning electric
  vehicle charging infrastructure,'' \emph{Transportation Research Part B:
  Methodological}, vol.~57, pp. 172--190, 2013.

\bibitem{sweda2011agent}
T.~Sweda and D.~Klabjan, ``An agent-based decision support system for electric
  vehicle charging infrastructure deployment,'' in \emph{2011 IEEE Vehicle
  Power and Propulsion Conference}.\hskip 1em plus 0.5em minus 0.4em\relax
  IEEE, 2011, pp. 1--5.

\bibitem{national2013transitions}
N.~R. Council \emph{et~al.}, \emph{Transitions to alternative vehicles and
  fuels}.\hskip 1em plus 0.5em minus 0.4em\relax National Academies Press,
  2013.

\bibitem{shareef2016review}
H.~Shareef, M.~M. Islam, and A.~Mohamed, ``A review of the stage-of-the-art
  charging technologies, placement methodologies, and impacts of electric
  vehicles,'' \emph{Renewable and Sustainable Energy Reviews}, vol.~64, pp.
  403--420, 2016.

\bibitem{zhang2015integrated}
H.~Zhang, Z.~Hu, Z.~Xu, and Y.~Song, ``An integrated planning framework for
  different types of pev charging facilities in urban area,'' \emph{IEEE
  Transactions on Smart Grid}, vol.~7, no.~5, pp. 2273--2284, 2015.

\bibitem{wang2013traffic}
G.~Wang, Z.~Xu, F.~Wen, and K.~P. Wong, ``Traffic-constrained multiobjective
  planning of electric-vehicle charging stations,'' \emph{IEEE Transactions on
  Power Delivery}, vol.~28, no.~4, pp. 2363--2372, 2013.

\bibitem{zhang2018joint}
H.~Zhang, S.~J. Moura, Z.~Hu, W.~Qi, and Y.~Song, ``Joint pev charging network
  and distributed pv generation planning based on accelerated generalized
  benders decomposition,'' \emph{IEEE transactions on transportation
  electrification}, vol.~4, no.~3, pp. 789--803, 2018.

\bibitem{wang_predictive}
B.~{Wang}, Y.~{Wang}, H.~{Nazaripouya}, C.~{Qiu}, C.~{Chu}, and R.~{Gadh},
  ``Predictive scheduling framework for electric vehicles with uncertainties of
  user behaviors,'' \emph{IEEE Internet of Things Journal}, vol.~4, no.~1, pp.
  52--63, 2017.

\bibitem{WANG20171289}
\BIBentryALTinterwordspacing
Y.~Wang, W.~Shi, B.~Wang, C.-C. Chu, and R.~Gadh, ``Optimal operation of
  stationary and mobile batteries in distribution grids,'' \emph{Applied
  Energy}, vol. 190, pp. 1289 -- 1301, 2017. [Online]. Available:
  \url{http://www.sciencedirect.com/science/article/pii/S0306261916319225}
\BIBentrySTDinterwordspacing

\bibitem{wencong_7763884}
T.~{Chen}, B.~{Zhang}, H.~{Pourbabak}, A.~{Kavousi-Fard}, and W.~{Su},
  ``Optimal routing and charging of an electric vehicle fleet for
  high-efficiency dynamic transit systems,'' \emph{IEEE Transactions on Smart
  Grid}, vol.~9, no.~4, pp. 3563--3572, 2018.

\bibitem{yang2020real}
B.~Yang, L.~Guo, and J.~Ye, ``Real-time simulation of electric vehicle
  powertrain: hardware-in-the-loop (hil) testbed for cyber-physical security,''
  in \emph{2020 IEEE Transportation Electrification Conference \& Expo
  (ITEC)}.\hskip 1em plus 0.5em minus 0.4em\relax IEEE, 2020, pp. 63--68.

\bibitem{bakirtzis2002optimal}
A.~G. Bakirtzis, P.~N. Biskas, C.~E. Zoumas, and V.~Petridis, ``Optimal power
  flow by enhanced genetic algorithm,'' \emph{IEEE Transactions on power
  Systems}, vol.~17, no.~2, pp. 229--236, 2002.

\bibitem{pencheva2009modelling}
T.~Pencheva, K.~Atanassov, and A.~Shannon, ``Modelling of a stochastic
  universal sampling selection operator in genetic algorithms using generalized
  nets,'' in \emph{Proceedings of the tenth international workshop on
  generalized nets, Sofia}, 2009, pp. 1--7.

\bibitem{deep2009real}
K.~Deep, K.~P. Singh, M.~L. Kansal, and C.~Mohan, ``A real coded genetic
  algorithm for solving integer and mixed integer optimization problems,''
  \emph{Applied Mathematics and Computation}, vol. 212, no.~2, pp. 505--518,
  2009.

\bibitem{deep2007new}
K.~Deep and M.~Thakur, ``A new mutation operator for real coded genetic
  algorithms,'' \emph{Applied mathematics and Computation}, vol. 193, no.~1,
  pp. 211--230, 2007.

\end{thebibliography}

\end{document}